# Weakly Supervised Pretraining and Multi-Annotator Supervised Finetuning for Facial Wrinkle Detection

Authors:

Ik Jun Moon[1†], Junho Moon[2†], Ikbeom Jang[3]*

Affiliations:

[1]Department of Dermatology, Asan Medical Center, University of Ulsan College of Medicine, Seoul, South Korea

[2]Department of Artificial Intelligence Semiconductor, Hanyang University, Seoul, South Korea

[3]Division of Computer Engineering, Hankuk University of Foreign Studies, Yongin, South Korea

Presenting author:

Full name: Ik Jun Moon

Affiliation: Department of Dermatology, Asan Medical Center, University of Ulsan College of Medicine, Seoul, South Korea

Email: ikjun.moon@amc.seoul.kr

Key information:

1. Research question: With the growing interest in skin diseases and skin aesthetics, the ability to predict facial wrinkles is becoming increasingly important. This study aims to evaluate whether a computational model, convolutional neural networks (CNN), can be trained for automated facial wrinkle segmentation.
2. Findings: Our study presents an effective technique for integrating data from multiple annotators and illustrates that transfer learning can enhance performance, resulting in dependable segmentation of facial wrinkles.
3. Meaning: This approach automates intricate and time-consuming tasks of wrinkle analysis with a deep learning framework. It could be used to facilitate skin treatments and diagnostics.

†: equal contribution

*: corresponding author

MANUSCRIPT

Introduction

With the growing interest in skin diseases and skin aesthetics, the ability to predict facial wrinkles is becoming increasingly important. Facial wrinkles are significant indicators of aging[1] and can be useful in skin conditions assessment[2], skin care, and early diagnosis of skin diseases[3]. We propose a deep learning-based approach to automatically segment facial wrinkles.

Analyzing extensive collections of images can be exceedingly resource-intensive if each facial wrinkle must be individually assessed. Moreover, the subjectivity inherent in manual segmentation processes can diminish the reliability of research findings and pose a substantial issue.

To address this issue, we effectively combine wrinkle data labeled by multiple annotators to minimize inter-rater variability and utilize these image-label pairs for training our model. Additionally, we enhanced performance using transfer learning that leverages the knowledge from a pretrained model. Unlike traditional methods, our approach utilizes knowledge from a pretrained model through transfer learning for downstream tasks, allowing it to achieve high performance even with limited labeled data. Furthermore, this method significantly reduces the time and cost involved in manually labeling wrinkles, offering substantial advantages over manual methods.

Material and methods

We utilized the FFHQ(Flickr-Faces-HQ)[4] dataset, which consists of 70,000 high-quality face images captured under various conditions. The images are 1024x1024 in size, and we used them without any downsampling or preprocessing. For pretraining, we randomly selected 25,000 images from this dataset. To generate the ground truth for weakly supervised pretraining, we extracted texture maps[5] from the face images and masked out non-facial areas to produce the final texture masks for ground truth (Figure 1-a). For finetuning, 500 face images were randomly selected. The ground truth for supervised finetuning consisted of manually annotated wrinkle masks. The wrinkle annotation process involved three annotators, all highly experienced in image processing and analysis. Recognizing the subjective nature of wrinkle identification, a consistent standard was established prior to the commencement of labeling. During the labeling, specific emphasis was placed on annotating wrinkles in key facial areas including the forehead, crow's feet, and nasolabial folds. To mitigate inter-rater variability and enhance the reliability of the ground truth, a majority voting mechanism was implemented, where only pixels labeled by at least two groups were retained (Figure 1-b). We allocated 80% of the dataset for training, 10% for validation, and 10% for testing.

Figure 2 shows the entire training pipeline. Initially, we trained the segmentation network in a weakly supervised manner using texture masks and subsequently finetuned it in a supervised manner using manually labeled masks. The training was performed using U-Net[6] architecture. During the weakly supervised pretraining stage, the model learns to output texture masks from face images, receiving images as input and producing texture masks as output (Figure 2-a). Then, in the supervised finetuning stage, the model learned to classify each pixel of a face image as either a wrinkle or not, using images and texture maps as inputs and producing two output classes that generate logit values for background and wrinkles, respectively (Figure 2-b).

In the weakly supervised pretraining stage, the model was trained for 300 epochs. Subsequently, supervised finetuning was performed for 150 epochs using manually annotated data. To evaluate the effectiveness of transfer learning in scenarios with limited label data, the supervised finetuning was performed using varying proportions of the whole training dataset: 100%, 50%, 25% and 5%. We used the Jaccard similarity index (JSI) as the evaluation metric, which is defined as follows:

$$JSI = \frac{|A \cap B|}{|A \cup B|}$$

where $A$ is the predicted segmentation, and $B$ is the actual label. This is a suitable metric for evaluating wrinkle segmentation performance as it measures the overlap between the predicted segmentation and the actual label.

Results

Table 1 displays JSI of our method compared to training exclusively with manual data (no pretraining). Our method demonstrates improved performance by 1.92%, 0.35%, 1.06%, and 8.53% for datasets comprising 100%, 50%, 25%, and 5% of the data, respectively. Additionally, our method demonstrates a significant performance improvement over methods pretrained with self-supervised learning techniques (Table 2). This highlights the efficacy of our proposed method in enhancing the accuracy of wrinkle segmentation, particularly in scenarios with sparse data availability. Figure 3 provides a visual performance of our methodology.

Discussion and Conclusion

We propose a reliable ground truth generation strategy and an efficient transfer learning approach. During the weakly supervised pretraining stage, the model learns to highlight skin features such as edges and textures. In the supervised finetuning stage, the model develops the capability to accurately discern wrinkle regions from edges and textures. Through this effective learning strategy, we can enhance the efficiency of wrinkle segmentation even with limited data, thereby achieving reliable wrinkle segmentation outcomes. However, despite majority voting, the subjectivity in wrinkle annotation remains a challenge. We plan to involve dermatologists in the wrinkle annotation process and research reliable label combination methods, such as label smoothing, to improve the reliability of manual wrinkle masks.

Disclosures

There is no disclosure to report.

APPENDIX

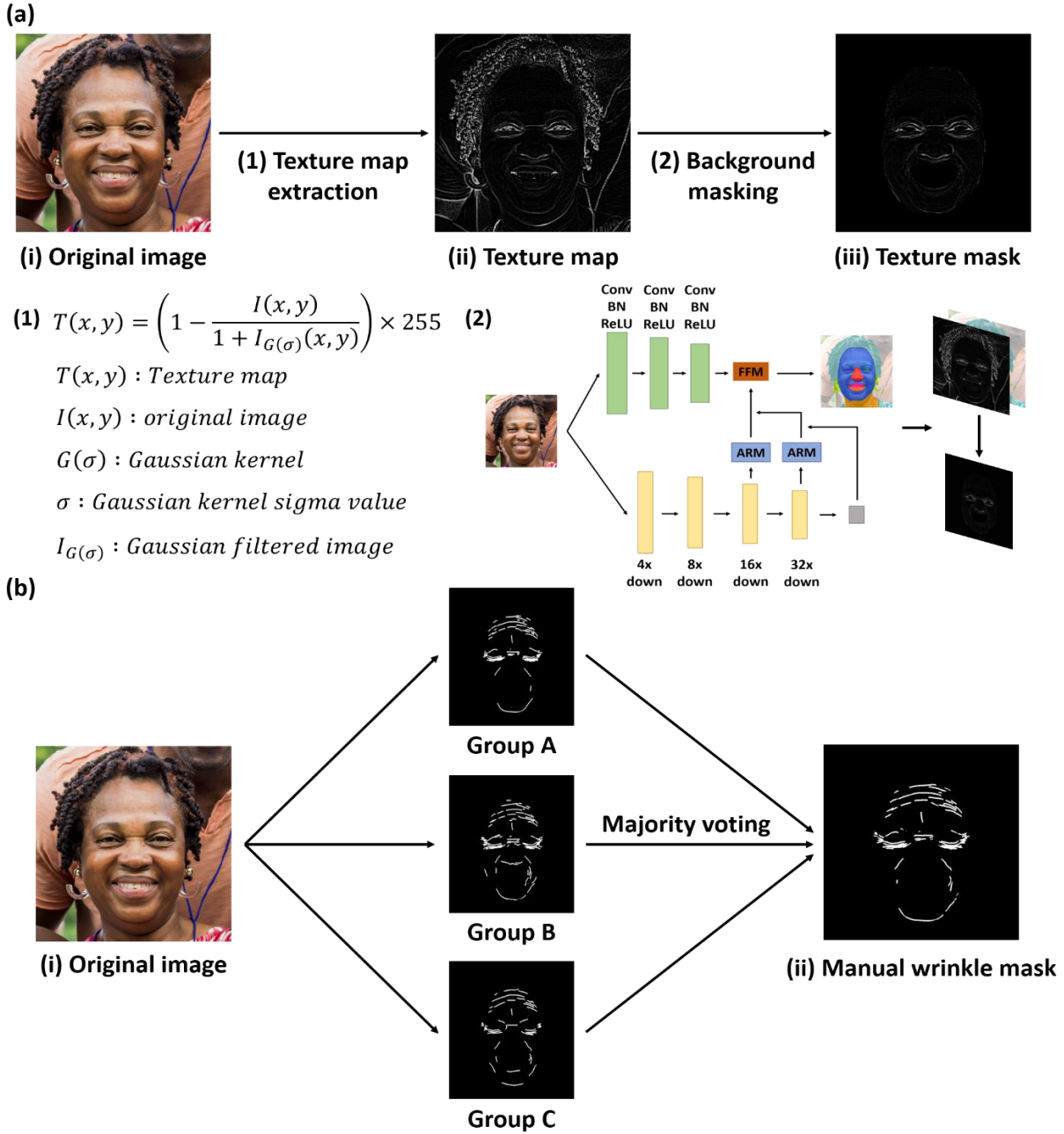

**Figure 1:** Ground truth wrinkle generation pipeline. (a) denotes the generation process of ground truth used during the weakly supervised pretraining stage. A texture map is extracted from the face image using a filter based on a Gaussian kernel. Since these texture maps contain numerous false positives from backgrounds, non-facial areas are subsequently masked using a face-parsing deep learning model based on the BiSeNet[7] architecture to produce the final texture mask. (b) denotes the ground truth generation process for the supervised finetuning stage. Three annotators each annotate the face images, and through majority voting, these annotations are combined to create the final manual wrinkle mask.

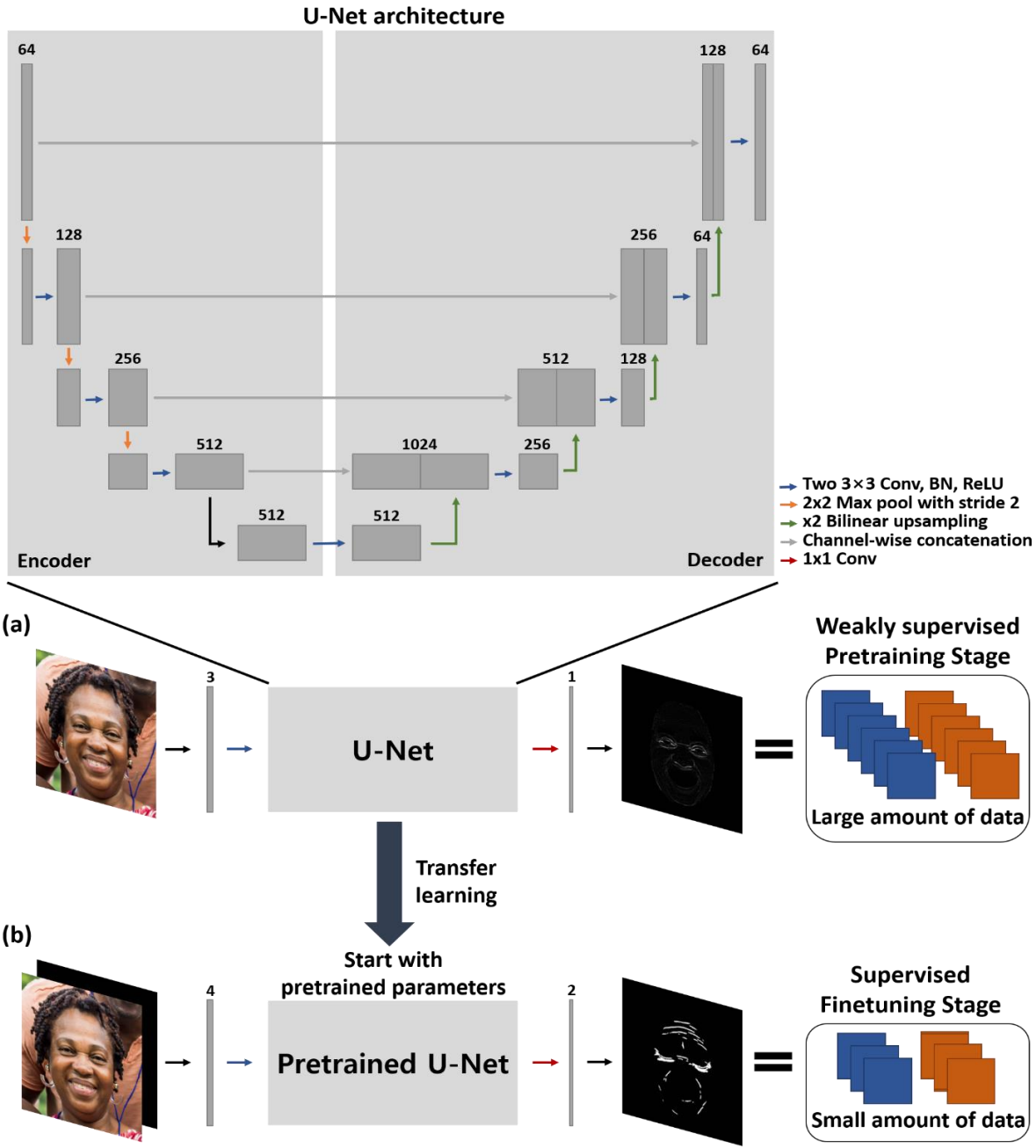

**Figure 2:** Weakly supervised pretraining and supervised finetuning with multi-annotator labels for facial wrinkle segmentation. (a) denotes the weakly supervised pretraining stage, where the segmentation network learns to extract texture masks from RGB face images. The model receives 3-channel RGB face images as input and produces 1-channel texture masks as output. (b) denotes the supervised finetuning stage, during which the segmentation network learns to extract facial wrinkles from RGB face images and texture masks. The model receives 3-channel RGB face images and 1-channel texture masks as input and produces 2-channel output classes, each generating logit values for background and wrinkles, respectively. By finetuning the weights of the model, which was trained with generic face texture extraction capabilities, we specifically enhanced its ability to detect facial wrinkles using manual data.

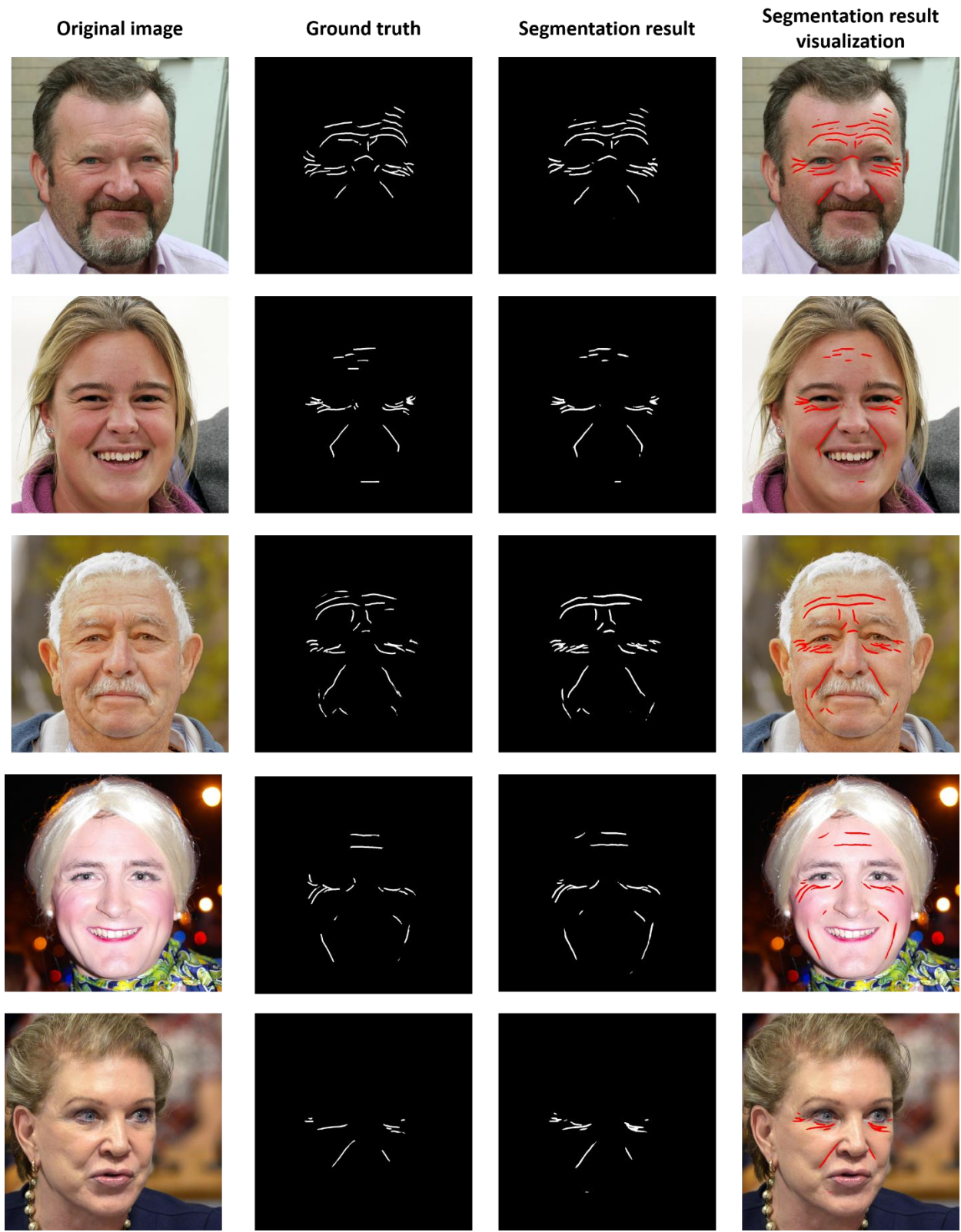

**Figure 3:** Test image samples and their predicted wrinkles compared with ground truth annotation.

**Table 1:** Segmentation performance comparison between a model trained exclusively with manually labeled wrinkle data (No pretraining) and our proposed method, which utilizes weakly supervised pretraining on filtered data followed by transfer learning. In the supervised finetuning stage, each method was trained on 100%, 50%, 25%, and 5% of the training dataset.

| Method | 100% (400) | 50% (200) | 25% (100) | 5% (20) | $n_{params}$ |
|---|---|---|---|---|---|
| No pretraining | 0.4319 | 0.4120 | 0.3594 | 0.2608 | 17.263M |
| Ours | **0.4511** | **0.4155** | **0.3700** | **0.3461** | 17.264M |

(#): number of datasets

**Table 2:** Model performance by different pretraining strategies. Jaccard similarity index illustrates the performance of models pre-trained using various self-supervised learning techniques compared to our proposed method. Our approach significantly outperforms other methods, demonstrating that our pretraining strategy is highly suitable for downstream tasks in wrinkle detection.

| Method | Reconstruction | Deblur | Denoise | Super-Resolution | Ours |
|---|---|---|---|---|---|
| Jaccard similarity index | 0.4334 | 0.4393 | 0.4364 | 0.4384 | **0.4511** |